\makeatletter\def\graphicscache@inhibit{true}\makeatother

\documentclass[letterpaper, 10 pt, conference]{ieeeconf}  %

\IEEEoverridecommandlockouts                              %

\usepackage[utf8]{inputenc}

\usepackage[hidelinks]{hyperref}
\usepackage[style=ieee,hyperref,natbib=true,backend=bibtex,firstinits,doi=false,%
     mincitenames=1,maxcitenames=2,maxbibnames=99,sorting=none,terseinits=false,hyperref=true]{biblatex}
\bibliography{references.bib}
\renewbibmacro*{bbx:savehash}{}%
\defbibheading{bibliography}[\bibname]{\section*{References}}

\usepackage{url}
\usepackage{graphicx}
\usepackage[usenames,dvipsnames]{xcolor}
\usepackage[draft]{fixme}
\fxsetup{theme=color}

\usepackage{amsmath}
\usepackage{amssymb}
\usepackage{textcomp}
\usepackage{siunitx}

\usepackage{booktabs}
\usepackage{threeparttable}
\usepackage{multirow}

\usepackage{calc}

\usepackage[capitalize]{cleveref}

\usepackage{tikz}
\usetikzlibrary{arrows}
\usetikzlibrary{positioning,calc}
\usetikzlibrary{decorations.pathreplacing}
\usetikzlibrary{decorations.markings}
\usetikzlibrary{fit}
\usetikzlibrary{shapes.callouts}
\usetikzlibrary{shapes.geometric}
\usetikzlibrary{matrix}

\usepackage{pgfplots}
\pgfplotsset{compat=1.9}
\usepgfplotslibrary{groupplots}
\usepgfplotslibrary{units}
\usepgfplotslibrary{statistics}
\usepgfplotslibrary{colorbrewer}
\usepackage{pgfplotstable}

\usepackage{letltxmacro,xcolor,xparse}
\usepackage[export]{adjustbox}

\usepackage{ifthen}
\IfFileExists{graphicscache.sty}{%
  \usepackage[dpi=440]{graphicscache}%
  \newcommand{\nocompress}[2][]{\includegraphics[##1,compress=false]{##2}}%
}{%
  \newcommand{\nocompress}[2][]{\includegraphics[##1]{##2}}%
}

\definecolor{teaserbg}{HTML}{ebebeb}

\usepackage{dblfloatfix}

\tikzset{
  font=\sffamily\footnotesize,
  m/.style={draw, rounded corners, fill=yellow!20, align=center}
}

\pgfplotsset{
        compat=1.7,
    }

\definecolor{t1}{RGB}{235, 172, 35}
\definecolor{t2}{RGB}{184, 0, 88}
\definecolor{t3}{RGB}{0, 140, 249}
\definecolor{t4}{RGB}{0, 110, 0}
\definecolor{t5}{RGB}{0, 187, 173}
\definecolor{t6}{RGB}{209, 99, 230}
\definecolor{t7}{RGB}{178, 69, 2}
\definecolor{t8}{RGB}{255, 146, 135}
\definecolor{t9}{RGB}{89, 84, 214}
\definecolor{t10}{RGB}{135, 133, 0}
\definecolor{t11}{RGB}{0, 198, 248}
\definecolor{t12}{RGB}{0, 167, 108}
\definecolor{t13}{RGB}{189, 189, 189}

\tikzdeclarecoordinatesystem{rel}{%
	\tikzset{cs/.cd,x=0pt,y=0pt,#1}%
	\pgfpointlineattime{(\tikz@cs@x)}%
	{\pgfpointanchor{\tikz@pp@name{\tikz@cs@node}}{south west}}%
	{\pgfpointanchor{\tikz@pp@name{\tikz@cs@node}}{south east}}%
	\edef\tikz@cs@x{\the\pgf@x}%
	\pgfpointlineattime{(\tikz@cs@y)}%
	{\pgfpointanchor{\tikz@pp@name{\tikz@cs@node}}{south west}}%
	{\pgfpointanchor{\tikz@pp@name{\tikz@cs@node}}{north west}}%
	\pgfpoint{\tikz@cs@x}{\pgf@y}%
}

\newcommand\currentcoordinate{\the\tikz@lastxsaved,\the\tikz@lastysaved}

\title{\LARGE \bf
Robust Immersive Telepresence and Mobile Telemanipulation:\\
NimbRo wins ANA Avatar XPRIZE Finals
}

\author{Max Schwarz$^{*}$, Christian Lenz$^{*}$, Raphael Memmesheimer, Bastian P\"atzold,\\ Andre Rochow, Michael Schreiber, and Sven Behnke%
\thanks{$^{*}$Equal contribution.}\thanks{All authors are with the Autonomous Intelligent Systems group of University of Bonn, Germany; {\tt schwarz@ais.uni-bonn.de}}%
}

\begin{document}

\maketitle

\begin{abstract}

Robotic avatar systems promise to bridge distances and reduce the need for
travel.
We present the updated NimbRo avatar system, winner of the \$5M grand prize
at the international ANA Avatar XPRIZE competition, which required
participants to build intuitive and immersive robotic telepresence systems
that could be operated by briefly trained operators.
We describe key improvements for the finals, compared to the system used in the semifinals:
To operate without a power- and communications tether, we integrated a battery
and a robust redundant wireless communication system.
Video and audio data are compressed using low-latency HEVC and Opus codecs.
We propose a new locomotion control device with tunable resistance force.
To increase flexibility, the robot's upper-body height can be adjusted by the operator.
We describe essential monitoring and robustness tools which enabled
the success at the competition.
Finally, we analyze our performance at the competition finals and discuss
lessons learned.

\end{abstract}

\section{Introduction}

Traveling large distances costs money and time; and most forms of travel impact the environment.
Reducing the need to travel is thus beneficial for many reasons.
While voice calls and video conferencing help, they cannot replace in-person meetings entirely
due to lack of immersion and social interaction.
Furthermore, many remote tasks require mobility, physical touch, grasping and handling of objects,
or even more complex manipulation skills.
These requirements cannot be addressed by VR-based conferencing systems that focus on meetings in a virtual space.
In contrast, avatar systems allow full immersion into a remote space while also \textit{embodying} the
operator in a robotic system, giving them the ability to navigate and physically interact with both the remote environment
and persons therein.
\begin{tikzpicture}[remember picture,overlay]
  \node[anchor=north,align=center,font=\sffamily\small,yshift=-0.4cm] at (current page.north) {%
  \textbf{Accepted final version.} IEEE-RAS International Conference on Humanoid Robots (HUMANOIDS), Austin, USA, December 2023};
\end{tikzpicture}%

The ANA Avatar XPRIZE competition\footnote{\url{https://www.xprize.org/prizes/avatar}} challenged the robotics community
to advance the state of the art in avatar systems. Promising a record \$10M prize purse, the competition
required teams to build \textit{intuitive} and \textit{robust} robotic avatar systems that allow a human operator
to be present in a remote space. The tasks to be solved included
social interaction and communication, but also locomotion and complex manipulation.
Critically, the systems were to be used and evaluated by operator and recipient judges.
In contrast to previous teleoperation competitions, such as the DARPA Robotics Challenge~\citep{krotkov2018darpa} operators
could be trained only for a short time how to use the developed avatar systems.

\begin{figure}
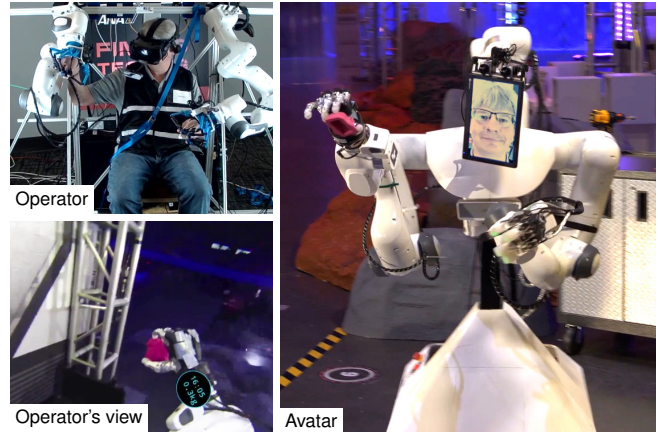

 \centering
 \newlength\teaserheight\setlength{\teaserheight}{5.6cm+3pt}
 \tikz[every node/.style={inner sep=0pt, outer sep=0pt}, lab/.style={inner sep=2pt,fill=white,font=\sffamily\scriptsize}]{
  \node (a) {\includegraphics[height=2.8cm,clip,trim=100px 650px 1360px 60px]{figures/jerry_stone_vr_contrast.jpg}};
  \node[below=3pt of a] (b) {\includegraphics[height=2.8cm,clip,trim=130px 250px 1330px 460px]{figures/jerry_stone_vr3_contrast.png}};
  \node[anchor=north west,xshift=3pt] at (a.north east) (c) {\includegraphics[clip,trim=750px 0px 400px 200px,height=\teaserheight]{figures/jerry_stone_arena_contrast.jpg}};
  
  \node[lab,anchor=south west,shift={(-0.1pt,-0.2pt)}] at (a.south west) {Operator};
  \node[lab,anchor=south west,shift={(-0.1pt,-0.2pt)}] at (b.south west) {Operator's view};
  \node[lab,anchor=south west,shift={(-0.1pt,-0.2pt)}] at (c.south west) {Avatar};
 }
 \caption{NimbRo avatar system at the ANA Avatar XPRIZE competition Finals. Stills from the end of our winning final run
 with the robot holding the correctly retrieved stone (magenta).
 Top left: Operator judge controlling the avatar.
 Bottom left: VR view (cropped).
 Right: Avatar robot in the arena.}
 \label{fig:teaser}
\end{figure}

In this paper, we present and discuss the updates and extensions of the NimbRo avatar system (\cref{fig:teaser})
that we made for our highly successful participation in
the ANA Avatar XPRIZE Finals in November 2022, where our team won the grand prize\footnote{\url{https://www.youtube.com/watch?v=EmESa2Olq4c}}.
The finals posed new requirements and tasks that resulted in various system extensions and improvements, compared to our earlier system used in the semifinals \mbox{\cite{schwarz2021nimbro}}.
The contributions of this paper include:

\begin{enumerate}
 \item hardware integration for tetherless and battery-powered operation of the avatar robot and mobility of the operator station,
 \item a redundant network stack for robust wireless communication,
 \item monitoring tools for efficient support crew operations,
 \item auto-recovery mechanisms for failure tolerance on multiple levels, and
 \item a thorough analysis of the competition results and lessons learned from our participation at the finals.
\end{enumerate}

\begin{figure*}
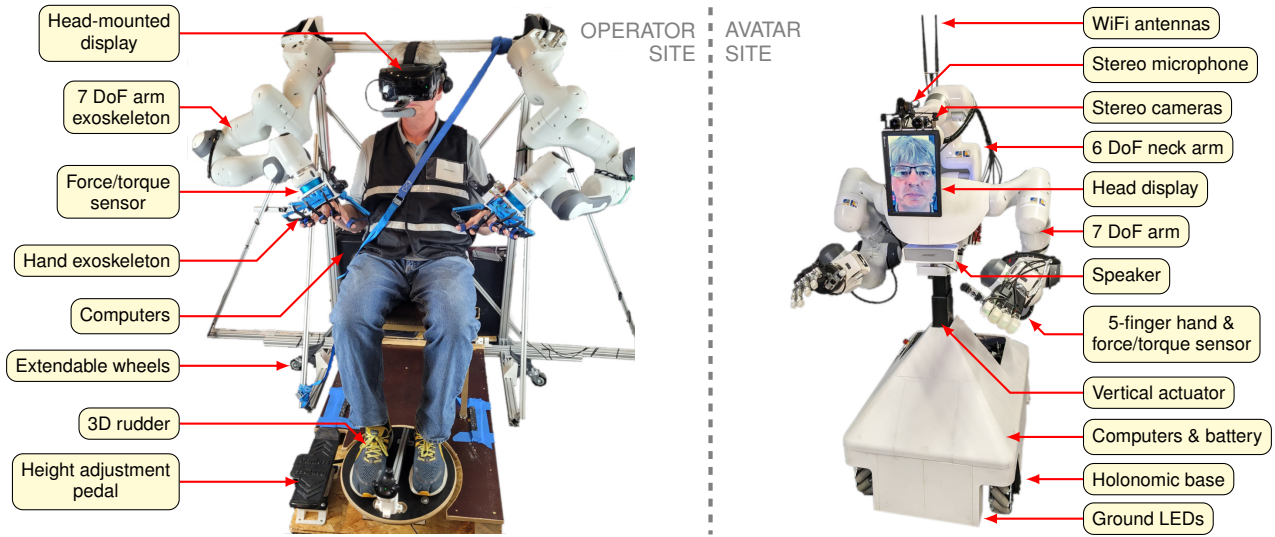

 \tikzset{
    n/.style={fill=yellow!20,rounded corners,draw=black,text=black,align=center,thin,font=\sffamily\scriptsize},
    lab/.style={draw=red,latex-,thick}
 }
 \centering
 \tikz[]{
  \node[inner sep=0pt] (otto) {\includegraphics[height=7cm]{figures/jerry.jpg}};

  \coordinate (ottow) at ($(otto.west)+(-0.1,0)$);

  \draw[lab] (rel cs:x=0.48,y=0.88,name=otto) -- ++(135:0.7cm) -- (\currentcoordinate-|ottow) node [n,anchor=east] {Head-mounted\\display};
  \draw[lab] (rel cs:x=0.1,y=0.77,name=otto) -- ++(135:0.4cm) -- (\currentcoordinate-|ottow) node [n,anchor=east] {7 DoF arm\\exoskeleton};
  \draw[lab] (rel cs:x=0.25,y=0.65,name=otto) --  (\currentcoordinate-|ottow) node [n,anchor=east] {Force/torque\\sensor};
  \draw[lab] (rel cs:x=0.25,y=0.6,name=otto) -- ++(225:0.8cm) -- (\currentcoordinate-|ottow) node [n,anchor=east] {Hand exoskeleton};

  \draw[lab] (rel cs:x=0.35,y=0.535,name=otto) -- ++(225:1.2cm) -- (\currentcoordinate-|ottow) node [n,anchor=east] {Computers};

  \draw[lab] (rel cs:x=0.23,y=0.32,name=otto) -- (\currentcoordinate-|ottow) node [n,anchor=east] {Extendable wheels};

  \draw[lab] (rel cs:x=0.4,y=0.17,name=otto) -- ++(135:0.4cm) -- (\currentcoordinate-|ottow) node [n,anchor=east] {3D rudder};
  \draw[lab] (rel cs:x=0.25,y=0.1,name=otto) -- (\currentcoordinate-|ottow) node [n,anchor=east] {Height adjustment\\pedal};

  \node[inner sep=0pt,right=2cm of otto] (anna) {\includegraphics[height=7cm]{figures/anna.jpg}};

  \coordinate (annae) at ($(anna.east)+(0.2,0)$);

  \draw[lab] (rel cs:x=0.55,y=0.97,name=anna) -- (\currentcoordinate-|annae) node [n,anchor=west] {WiFi antennas};

  \draw[lab] (rel cs:x=0.45,y=0.82,name=anna) -- ++(45:0.7cm) -- (\currentcoordinate-|annae) node [n,anchor=west] {Stereo microphone};
  \draw[lab] (rel cs:x=0.53,y=0.79,name=anna) -- ++(45:0.2cm) -- (\currentcoordinate-|annae) node [n,anchor=west] {Stereo cameras};

  \draw[lab] (rel cs:x=0.7,y=0.735,name=anna) -- (\currentcoordinate-|annae) node [n,anchor=west] {6 DoF neck arm};

  \draw[lab] (rel cs:x=0.55,y=0.655,name=anna) -- (\currentcoordinate-|annae) node [n,anchor=west] {Head display};

  \draw[lab] (rel cs:x=0.9,y=0.575,name=anna) -- (\currentcoordinate-|annae) node [n,anchor=west] {7 DoF arm};

  \draw[lab] (rel cs:x=0.6,y=0.52,name=anna) -- ++ (-45:0.3cm) -- (\currentcoordinate-|annae) node [n,anchor=west] {Speaker};

  \draw[lab] (rel cs:x=0.86,y=0.41,name=anna) -- ++ (-45:0.3cm) -- (\currentcoordinate-|annae) node [n,anchor=west] {5-finger hand \&\\force/torque sensor};

  \draw[lab] (rel cs:x=0.55,y=0.40,name=anna) -- ++(-45:1.3cm) -- (\currentcoordinate-|annae) node [n,anchor=west] {Vertical actuator};

  \draw[lab] (rel cs:x=0.78,y=0.185,name=anna) -- (\currentcoordinate-|annae) node [n,anchor=west] {Computers \& battery};

  \draw[lab] (rel cs:x=0.85,y=0.105,name=anna) -- (\currentcoordinate-|annae) node [n,anchor=west] {Holonomic base};

  \draw[lab] (rel cs:x=0.7,y=0.03,name=anna) -- (\currentcoordinate-|annae) node [n,anchor=west] {Ground LEDs};

  \coordinate (split) at ($(otto.north east)!0.5!(anna.north west)$);
  \begin{scope}[draw=black!50,text=black!50]
    \node[below right=0.1cm of split,align=left] {AVATAR\\SITE};
    \node[below left=0.1cm of split,align=right] {OPERATOR\\SITE};
    \draw[dashed,ultra thick] (split) -- ($(otto.south east)!0.5!(anna.south west)$);
  \end{scope}
 }
 \caption{NimbRo avatar system consisting of the operator station (left) and the avatar robot (right).}
 \label{fig:sys_overview}
\end{figure*}

\section{Related Work}

Teleoperated robotic systems are widespread. For example, the DARPA Robotics Challenge~\citep{krotkov2018darpa}
resulted in an array of legged, wheeled, and tracked teleoperated robots.
We focus our discussion on avatar systems, that is, systems which not only
allow teleoperation, but also telepresence---immersion in the remote environment and interaction with human recipients, as in the XPRIZE competition.

Some participants are commercial entities and have no scientific publications, such as
Pollen Robotics, who came in second in the finals.
Their avatar robot is a heavily modified version of their product
\textit{Reachy} with stronger arms (7\,DoF, 3.5\,kg payload) and a
communication head displaying the operator's face.
In contrast to our operator station, Pollen uses VR controllers with vibration actuators and a 1\,DoF elbow exoskeleton for manipulation.
Torque and haptic feedback is thus limited to elbow torque and controller vibration.

\Citet{luo2023northeastern} describe the approach by team Northeastern, who achieved third place at the finals.
Their avatar system features an interesting glove and gripper system with hydrostatic actuation, which
gives fine-grained force feedback. A wave variable approach handles varying communication latency.
The decision to forgo a VR head-mounted display in favor of 2D screens is a simple way to avoid
motion sickness caused by network latency, but limits immersiveness in comparison to our system.

\Citet{marques2022commodity} came in fourth with the AVATRINA system.
Mechanically, it is similar to our system with two Franka Emika Panda arms, dexterous hands and
VR teleoperation. In contrast to ours, the neck can only rotate and not translate,
which makes looking around occlusions impossible and reduces depth perception and immersion.

\Citet{van2022comes} showcase the system of team i-Botics, who came in fifth.
Based on Halodi Eve, their robot is considerably more humanoid in shape than the other top five teams,
although it operated with a gantry for safety reasons up until the last competition run.
During the competition, the system suffered from network connectivity problems.
The robot's neck is only 1\,DoF, seriously limiting camera motion.

\Citet{park2022team} describe the system of team SNU ($8^{\textrm{th}}$ place).
Their avatar robot is fully humanoid in shape, although for finals it was sitting on a holonomic base
for fast and safe locomotion.
Interestingly, the team decided not to attempt realistic animation of the operator's face,
but displayed operator emotions through basic line drawings of the mouth area,
which limits identification of the robot with the human controlling it.
The SNU operator station features a unique linear actuator system which seems to
enlarge the workspace.

Aside from the XPRIZE competition, there are other notable avatar systems targeting different applications like enabling people with disabilities to work remotely~\citep{takeuchi2020avatar}, underwater telepresence~\citep{khatib2016ocean}, and space~\citep{lii2017toward}.
\Citet{takeuchi2020avatar} presented an avatar robot system with focus on enabling people with disabilities to execute physical work remotely. 
The robot is controlled by mouse and gaze input.
In contrast to our proposed system, which aims at transmitting the movements of an operator naturally, their system follows a less immersive approach by controlling motion through a GUI.
\Citet{lii2017toward} describe a teleoperated robot controlled by astronauts
through a tablet GUI and a 1-DoF haptic joystick.
The system can execute assembly and maintenance tasks but has, due to the domain,
no human interaction capabilities.
TELESAR VI~\citep{tachi2020telesar} has a long history in the  field of telexistence, starting in 1980. The system is designed for remote manipulation and gesturing.
Unlike ours, the robot is operated in a stationary seated posture and has two controllable legs. All ten fingers are equipped with multiple sensors for enhanced tactile feedback.
However, the force feedback is limited to the fingers.

\section{NimbRo Avatar System}

The NimbRo avatar system consists of a robotic operator station and an anthropomorphic avatar robot (see \cref{fig:sys_overview}).
The human operator sits on a chair and is strapped into two 7\,DoF compliant arm exoskeletons (Franka Emika Panda arms) at their palms.
6\,DoF force/torque sensors (Nordbo NRS-6050-D80) at the arm wrists are used to provide a weightless feeling for the operator and force/torque feedback to their hands.
Finger movements are captured using SenseGlove DK1 hand exoskeletons, which provide force and haptic feedback to the finger tips.
The operator's feet are resting on a custom-built 3D pedal device (see \cref{sec:locomotion_control}), which allows omnidirectional control of the robot's base.
For visual and auditive immersion, the operator wears a VR head-mounted display (Valve Index), which is equipped with additional cameras to capture gaze direction, eye opening, and mouth expressions.

The avatar robot is equipped with a holonomic mecanum-wheeled base for indoor locomotion.
Its spine is a linear actuator that can be used to adapt to different manipulation and communication heights.
For bimanual manipulation, the robot features two arms (Franka Emika Panda) in approximately humanoid configuration.
Force/torque sensors (OnRobot HEX-E) are mounted on the wrists for force feedback. We chose two different
robotic hands (Schunk SVH and SIH), each featuring different capabilities.
Please refer to \citet{lenz2023bimanual} for a detailed description of our arm force feedback telemanipulation controller,
including active joint limit avoidance using model-based predictions and an oscillation detection and suppression module.

The robot head is mounted on a robotic arm (UFactory xArm-6), which mirrors the operator's 6D head movement.
Together with the head-mounted wide-angle stereo camera pair (2$\times$Basler a2A3840-45ucBAS) this
enables a highly immersive visual 3D experience for the operator. Latencies are mitigated by spherical rendering~\citep{schwarz2021vr}.
The head further carries a screen showing a live animation of the operator's face---a direct video display is not
possible, since the operator is wearing the HMD. See~\citep{rochow2022vr,rochow2023attention} for details on the facial animation.

The first integrated NimbRo avatar system (as of February 2021, prior to the ANA Avatar XPRIZE Semifinals) is described in detail in \citet{schwarz2021nimbro}.

The ANA Avatar XPRIZE finals~\cite{XPRIZE2022Rules} brought new tasks and requirements which necessitated changes to our design.
On the one hand, it was clear that the robot had to operate without a tether for communications and power, which necessitated integration of wireless communication and battery power.
The robot had to navigate through narrow passages, necessitating to reduce its width.
Much emphasis was placed by XPRIZE on haptic perception and we extended our system accordingly.
On the other hand, the competition format changed: Whereas in the semifinals, individual scenarios could
be attempted multiple times, with manual intervention allowed in-between, the finals called for a continuous mission
through ten tasks, with no possibility to skip tasks or to re-start the system in case of failure.
Additionally, the participants were down-selected over the successive competition days (see \cref{sec:eval}).
This meant that considerable focus had to be placed on making the system robust.

We will now detail the changes and improvements to the system since early 2021 up to the finals in November 2022.

\subsection{Mobile Operator Station}

The finals required a mobile operator station that could be moved into an operator control room and set-up quickly. To this end,
we added extendable wheels to it (see \cref{fig:sys_overview}). The control computer and a dedicated computer for
face animation were integrated into the structure. Additionally, we added a large battery
which can supply the operator station for multiple hours.
Using this setup, we could prepare the operator station for use long before our run and leave everything
initialized and switched on during transport to the operator control room.

\subsection{Telemanipulation}

Our telemanipulation components, including the arm and hand exoskeletons on the operator side, arms and five-finger hands on the 
avatar side, and our force feedback controller have been proven during semifinals~\cite{lenz2023bimanual}.
To adapt to the finals requirements, we made three changes.
First, we mounted the avatar's arm bases closer together, which reduced the shoulder width and made it easier to maneuver through narrow passages.
Second, we equipped the fingertips of the SVH and SIH hands with microswitches and magnet hall sensors, respectively.
This allows contact to be measured and then displayed haptically to the operator using the SenseGlove DK1.
Finally, we replaced the OnRobot HEX\nobreakdash-E force/torque sensors on the operator side with more rigid Nordbo NRS-6050-D80 which offer
higher update rate of 1\,kHz, resulting in faster response to operator movement.

\subsection{Roughness Sensing}

Task~10 of the finals (see \cref{fig:tasks}) required the ability to remotely feel texture, especially surface roughness.
To enable this, we developed an audio-based sensing, detection, and vibrational display system:
The left index fingertip is equipped with two microphones: one measuring air vibrations, the other measuring vibrations coupled
through the finger. The audio data is transmitted over WiFi (see \cref{sec:wifi}) and then
analyzed by a CNN, which classifies very short audio segments as rough or smooth.
Finally, a vibration actuator on the operator's fingertip displays the classification result,
giving the illusion of feeling rough bumps on the object surface with low latency.
This low-cost approach is very small and non-invasive on the sensing side.
Details of this subsystem are described by \citet{paetzold2023haptics}.

\subsection{Tetherless Operation}

To make the system operable without a tether, wide-ranging changes to the avatar robot had to be implemented.

\subsubsection{Power Supply}

\begin{table}
 \centering \footnotesize
 \caption{Avatar Robot Power Distribution \& Average Consumption}\label{tab:power}
 \begin{tabular}{@{}r@{\hspace{2pt}}lr|r@{\hspace{2pt}}lr@{}}
  \toprule
  Voltage\hspace{.2cm} & Device & Power & Voltage\hspace{.2cm} & Device & Power \\
  \midrule
  \multirow{2}{*}{\tikz[baseline={(0,0)}]{\node[font=\rmfamily,anchor=east]{Battery}; \draw (0,0) |- +(0.1,0.2); \draw (0,0) |- +(0.1,-0.2);}}
                           & Wheels (idle) & 16\,W  & \multirow{3}{*}{\tikz[baseline={(0,0)}]{\node[font=\rmfamily,anchor=east]{24V}; \draw (0,0) |- +(0.1,0.3); \draw (0,0) |- +(0.1,-0.3);}}
                           & xArm motors   & 50\,W \\
                           
                           & PC            & 350\,W & & Hands         & 8\,W \\
  \tikz[baseline={(0,-0.08)}]{\node[font=\rmfamily,anchor=east]{5V}; \draw (0,0) -- (0.1,0.0);}
                           & Base controller & 5\,W & & F/T sensors   & 11\,W \\
  \multirow{2}{*}{\tikz[baseline={(0,0)}]{\node[font=\rmfamily,anchor=east]{12V}; \draw (0,0) |- +(0.1,0.2); \draw (0,0) |- +(0.1,-0.2);}}
                           & xArm PC       & 15\,W & \tikz[baseline={(0,-0.08)}]{\node[font=\rmfamily,anchor=east]{48V}; \draw (0,0) -- (0.1,0.0);}
                           & Panda motors  & 60\,W \\
                           & Panda PCs     & 145\,W \\
  \midrule
  \multicolumn{6}{c}{Total power consumption (avg): 660\,W} \\
  \bottomrule
 \end{tabular}
 \vspace{-1ex}
\end{table}

\begin{figure}
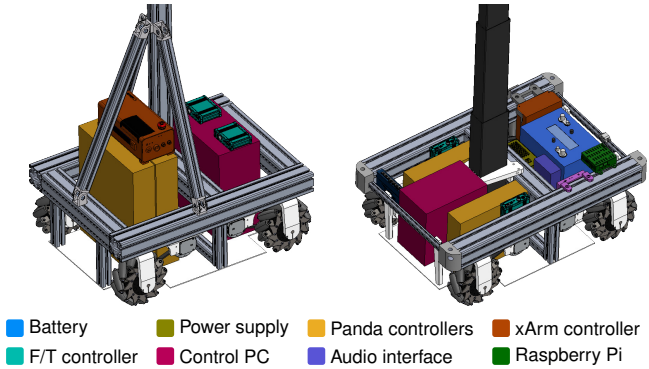

 \centering
 \hfill
 \includegraphics[height=4cm,clip,trim=0px 0px 0px 100px]{figures/Atlas2021.png}
 \hfill
 \includegraphics[height=4cm,clip,trim=0px 0px 0px 100px]{figures/Atlas2022.png}
 \hfill\strut
 \tikz[
     b/.style={minimum width=5pt,minimum height=5pt,rounded corners=1pt},
     font=\sffamily\scriptsize,
     every node/.style={outer sep=0pt},
     every label/.style={inner sep=2pt}
   ]{
   \node[matrix,column sep=5pt,row sep=0pt] {
     \node[b,fill=t3,label={east:Battery}] {}; &
     \node[b,fill=t10,label={east:Power supply}] {}; &
     \node[b,fill=t1,label={east:Panda controllers}] {}; &
     \node[b,fill=t7,label={east:xArm controller}] {}; \\
     \node[b,fill=t5,label={east:F/T controller}] {}; &
     \node[b,fill=t2,label={east:Control PC}] {}; &
     \node[b,fill=t9,label={east:Audio interface}] {}; &
     \node[b,fill=t4,label={east:Raspberry Pi}] {}; \\
   };
 }
 \vspace{-3.5ex}
 \caption{Robot base before (left) and after update (right).
 }\label{fig:cad_base}
\end{figure}

The robot now carries a RELiON InSight 48\,V 30\,Ah battery, which powers all onboard systems.
All power consumers had to be converted to DC power. We identified four voltage rails (see \cref{tab:power})
and installed individual DC-to-DC converters. This required tight integration to fit
everything into the robot base (see \cref{fig:cad_base}).

\subsubsection{Wireless Communication}\label{sec:wifi}

\begin{figure}
 \centering
 \begin{tikzpicture}[transmission/.style={decorate, decoration={expanding waves, angle=30,
                          segment length=3}}]
  \node[m,minimum width=1.6cm,minimum height=1.3cm,label={[align=center]north:Operator\\station PC}] (op) at (0,0) {};
  \node[anchor=east,draw,fill=orange!40] (eth) at (op.east) {Eth NIC};
  \node[anchor=south,rotate=90,font=\sffamily\tiny,inner sep=2pt] at (eth.west) {PCI-E};

  \node[m,fill=blue!20,minimum width=1.6cm,minimum height=1.3cm,right=1.3cm of op,label={[align=center]north:XPRIZE\\network}] (xprize) {};
  \node[anchor=east,draw,fill=orange!40] (ap5) at ($(xprize.east)+(0,+0.3)$) {AP};
  \node[anchor=east,draw,fill=orange!40] (ap2) at ($(xprize.east)+(0,-0.3)$) {AP};

  \node[m,minimum width=1.8cm,minimum height=1.3cm,right=2.2cm of xprize,label={[align=center]north:Avatar PC}] (avatar) {};
  \node[anchor=west,draw,fill=orange!40] (w5) at (ap5-|avatar.west) {WiFi NIC};
  \node[anchor=west,draw,fill=orange!40] (w2) at (ap2-|avatar.west) {WiFi NIC};
  \node[anchor=north,rotate=90,font=\sffamily\tiny,inner sep=2pt] at (w5.east) {PCI-E};
  \node[anchor=north,rotate=90,font=\sffamily\tiny,inner sep=2pt] at (w2.east) {PCI-E};

  \draw (op) -- (xprize) node[midway,align=center,font=\sffamily\scriptsize] {1\,GBit/s\\Ethernet};

  \node (wifi5) at ($(ap5.east)!0.5!(w5.west)$) {5\,GHz};
  \node (wifi2) at ($(ap2.east)!0.5!(w2.west)$) {2.4\,GHz};

  \draw[transmission] (ap5.east) -- ++(0.5,0);
  \draw[transmission] (w5.west) -- ++(-0.5,0);

  \draw[transmission] (ap2.east) -- ++(0.5,0);
  \draw[transmission] (w2.west) -- ++(-0.5,0);

  \draw (xprize.west) -- ++(0.3,0) |- (ap5);
  \draw (xprize.west) -- ++(0.3,0) |- (ap2);
 \end{tikzpicture}
 \caption{Network Architecture. The operator station contains a 1\,GBit/s ethernet adapter, which is connected
 to the XPRIZE network (or our own access point during testing). Two separate access points broadcast a WiFi network at
 2.4\,GHz and 5\,GHz, respectively.
 The avatar control PC is equipped with two PCI-E WiFi adapters, one for each of the networks.}
 \label{fig:network_architecture}
\end{figure}
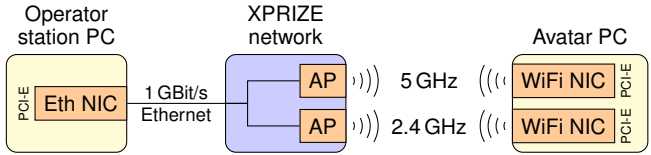

For the finals, XPRIZE supplied a 2.4\,GHz and a 5\,GHz WiFi network on the competition arena.
To utilize this fully, we incorporated two WiFi adapters on our robot (see \cref{fig:network_architecture}).
All network adapters (wired on the operator side, wireless on the robot side) are connected via PCI-E to
their respective control computers---reducing latency to a minimum and increasing robustness, compared to
USB devices.
On the robot, we use two Intel AX210 cards, which have excellent driver support. Each WiFi card is connected to two antennas: a main antenna extended up from
the spine of the robot (see \cref{fig:sys_overview}) and a backup antenna in the base.

To facilitate fine-grained control of routing, i.e. which WiFi band is used for which data stream,
each WiFi adapter has its own IP address. Conversely, the operator station uses two IP addresses on its
Ethernet interface. Static routes ensure that the correct source address and interface are used for
each destination address.

Our network transport is based on \texttt{nimbro\_network}\footnote{\url{https://github.com/AIS-Bonn/nimbro_network}},
which transmits data between the otherwise isolated ROS~\citep{Quigley09} instances.
The type and amount of data is statically configured and transmitted over UDP, which avoids any kind of connection handshake.
This allows our system to seamlessly begin operation as soon as network connectivity is established and to recover
immediately after a network interruption.
This capability strongly contributed to our success at the finals (see \cref{sec:eval}).

At semifinals in 2021, our system required around 300\,MBit/s downlink bandwidth. At the time, this was fine
given a wired 1\,GBit/s connection, but now bandwidth had to be reduced for robust WiFi operation.
While current WiFi systems can sustain such bandwidths in ideal situations, they cannot
guarantee this in non-line-of-sight or otherwise difficult circumstances.
We reduced the required bandwidth significantly by compressing the main video stream (2$\times$2472$\times$2178\,@\,46\,Hz)
with the HEVC video codec instead of transmitting individual JPEG images.
We took care not to increase video latency, which is a source of operator disorientation and fatigue.
To this end, we perform debayering, white balancing, and color correction in a fused CUDA kernel on the onboard
RTX 3070 GPU. Video encoding is then performed using the NVIDIA NVENC library on the GPU as well.
On the operator side, compressed video packets are uploaded to the GPU (RTX A6000 48\,GB), extracted using NVDEC,
and sent to the HMD after spherical rendering~\citep{schwarz2021vr}, which corrects for head movement latencies.
The total latency from start of camera exposure to HMD display is under 50\,ms.
Overall, we achieve considerably lower bandwidth than our old system (see \cref{tab:downlink}).

\begin{table}
 \centering\scriptsize\setlength{\tabcolsep}{4pt}
 \caption{WiFi Bandwidth Requirements}\label{tab:downlink}\label{tab:uplink}
 \begin{tabular}{@{}l@{\hspace{-0ex}}rc@{\hspace{2pt}}cl@{\hspace{-0ex}}rc@{\hspace{2pt}}c@{}}
  \toprule
  \multicolumn{4}{c}{Downlink from avatar} & \multicolumn{4}{c}{Uplink to avatar}\\
  \cmidrule (r) {1-4} \cmidrule (l) {5-8}
  Channel & MBit/s & 5\,GHz & 2.4\,GHz & Channel & MBit/s & 5\,GHz & 2.4\,GHz \\
  \cmidrule (r) {1-4} \cmidrule (l) {5-8}
  Arm feedback            &  8.5 & $\checkmark$ & $\times$     & Arm control             &  4.9 & $\checkmark$ & $\checkmark$ \\
  Transformations         &  4.1 & $\checkmark$ & $\times$     & Transformations         &  1.4 & $\checkmark$ & $\times$ \\
  Main cameras            & 14.7 & $\checkmark$ & $\times$     & Operator face  &  5.7 & $\times$     & $\checkmark$ \\
  Hand camera     &  5.5 & $\times$     & $\checkmark$ & Audio                   &  0.4 & $\checkmark$ & $\checkmark$ \\
  Diagnostics      &  0.4 & $\checkmark$ & $\checkmark$ \\
  Audio                   &  0.4 & $\checkmark$ & $\checkmark$ \\
  \cmidrule (r) {1-4} \cmidrule (l) {5-8}
  Total [MBit/s]                   &              & 28.1 & 6.3 & Total [MBit/s]            &              & 6.7 & 11.0 \\
  \bottomrule
 \end{tabular}
\end{table}

\Cref{tab:downlink} also shows that manipulation control and audio are routed over both WiFi networks
redundantly. Both data types are very sensitive to interruptions and packet drops, which lead to
uncontrolled movement and loss of information, respectively. The redundant transmission minimizes this risk. We also experimented with redundant
configuration for our camera stream, but refrained from activating this in the finals due to concerns about
2.4\,GHz bandwidth in an arena with many devices of spectators operating in this band.

Audio transmission over WiFi is in itself problematic. Our semifinal solution used the Opus codec with a very
low buffer size of 64 samples (at 48\,kHz audio), which gives a high packet rate---challenging for WiFi networks.
The packet rate can be reduced by increasing the buffer size, but this increases latency and easily leads to
echo effects, where the operator can hear their own voice as transmitted by the avatar and then captured by the
avatar's microphones. To mitigate this, we integrate an echo cancellation system based on NVIDIA Maxine\footnote{\url{https://developer.nvidia.com/maxine}}, which allows us to increase the buffer size to 512,
reducing the audio packet rate to roughly 90\,Hz.

\subsubsection{Wireless E-Stop}\label{sec:e-stop}

We integrated an HRI Wireless Emergency Stop to provide a reliable tetherless safety system.
Activating the E-Stop depowers the wheels.
The avatar stops quickly due to friction, and the base can be pushed around manually.
In addition, both Panda arms, the xArm, and both hands hold their current joint positions.
The Panda arms can be moved freely using the teach button on their wrist.

\subsection{Height Adaption}

In contrast to semifinals, where manipulation on table height was required, finals specified a wider height range.
To increase flexibility, we integrated a linear joint in the spine of the robot (see \cref{fig:sys_overview}).
The operator can control the height with a bidirectional Danfoss KEP foot pedal.
The pedal contains springs to provide resistance and uses hall effect sensors for angle sensing.
The current height and a side-view rendering of the robot is shown to the operator during movement of the actuator to assist the height adjustment.

\subsection{Locomotion Control}
\label{sec:locomotion_control}

\begin{figure}
  \tikzset{
    n/.style={fill=yellow!20,rounded corners,draw=black,text=black,align=center,thin,font=\sffamily\scriptsize,text depth=0pt,text height=4pt},
    lab/.style={draw=red,latex-,thick}
  }
  \centering
  \tikz[]{
    \node[inner sep=0pt] (rudder) {\includegraphics[height=3cm]{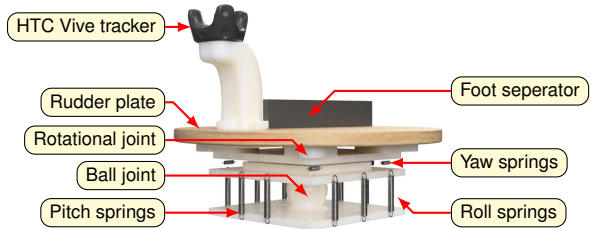}};

    \coordinate (rudderw) at ($(rudder.west)+(-0.1,0)$);

    \draw[lab] (rel cs:x=0.07,y=0.9,name=rudder) -- (\currentcoordinate-|rudderw) node [n,anchor=east] {HTC Vive tracker};
    \draw[lab] (rel cs:x=0.1,y=0.45,name=rudder) -- ++(135:0.5cm) -- (\currentcoordinate-|rudderw) node [n,anchor=east] {Rudder plate};
    \draw[lab] (rel cs:x=0.5,y=0.33,name=rudder) -- ++(135:0.3cm) -- (\currentcoordinate-|rudderw) node [n,anchor=east] {Rotational joint};
    \draw[lab] (rel cs:x=0.5,y=0.15,name=rudder) -- ++(135:0.4cm) -- (\currentcoordinate-|rudderw) node [n,anchor=east] {Ball joint};
    \draw[lab] (rel cs:x=0.25,y=0.08,name=rudder) -- (\currentcoordinate-|rudderw) node [n,anchor=east] {Pitch springs};

    \coordinate (ruddere) at ($(rudder.east)+(0.1,0)$);

    \draw[lab] (rel cs:x=0.5,y=0.5,name=rudder) -- ++(45:0.5cm) -- (\currentcoordinate-|ruddere) node [n,anchor=west] {Foot seperator};
    \draw[lab] (rel cs:x=0.8,y=0.3,name=rudder) -- (\currentcoordinate-|ruddere) node [n,anchor=west] {Yaw springs};
    \draw[lab] (rel cs:x=0.83,y=0.15,name=rudder) -- ++(-45:0.3cm) -- (\currentcoordinate-|ruddere) node [n,anchor=west] {Roll springs};
  }
  \caption{Self-centering rudder design with individually tunable springs for intuitive omnidirectional locomotion control. \cref{fig:sys_overview} shows the device during operation.}
  \vspace{-1ex}
  \label{fig:rudder}
\end{figure}

For locomotion of the holonomic robot platform, we propose a new feet-controlled 3D-printed rudder design (see \cref{fig:rudder}).
We identified two major problems with the old device: First, the rudder lacked resistance and self-centering, which made it difficult to control.
Second, the rudder yielded only position estimates when both feet were placed on the rudder surface. 
This became especially problematic when the feet were lifted off the rudder and tilted or rotated slightly, necessitating an often unintuitive, lengthy re-initialization step for operators.

To address these issues, we introduced resistance and self-centering by employing a spring mechanism.
The mechanical base of the rudder is built around a ball-bearing joint and a rotational thrust-bearing joint.
By employing two joints, we have control over the pitch, roll and yaw axes by using different springs.
Springs with different tension allow individual control of the resistance per axis.
For absolute position estimates, we attach an HTC Vive tracker to the rudder which receives signals from the VR tracking system that are then translated to movement commands.
We place a separator on the rudder's surface to ease blind foot placement.

We equipped the avatar robot with addressable LED strips (90 RGB LEDs) under the base, which indicate the driving direction by illuminating the corresponding side of the robot. The LEDs also show the battery level during charging.

\subsection{Monitoring}

\begin{figure}
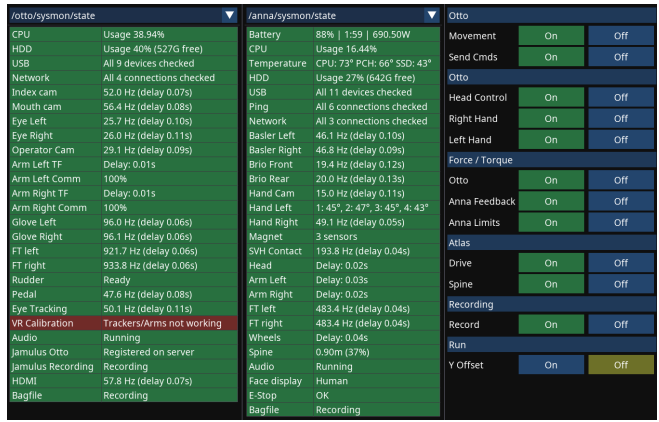

 \centering
 \nocompress[width=\linewidth,clip,trim=10px 10px 10px 10px]{figures/imgui_sysmon.png}
 \caption{System Monitoring GUI. Left: Operator Station status. Each line corresponds to a system check.
 The red check indicates an issue with the VR trackers mounted on the exoskeleton---caused by
 a support crew member occluding the line-of-sight.
 Center: Avatar robot status.
 Right: Control buttons that enable/disable individual system components.}
 \vspace{-1ex}
 \label{fig:sysmon}
\end{figure}

Monitoring is an essential part of robust robotics. It allows engineers to analyze problems and find their
causes quickly.
In our scenario, it was especially important to make sure the system is healthy before starting the run, since
from then on, manual intervention was not permitted. During the run, the role of monitoring switches to
a safety perspective, allowing the support crew to abort the run in case of danger to the human operator, the robot,
or the environment.
To be able to monitor the highly complex avatar system with one glance, we developed an integrated GUI.
Because it contains a multitude of video streams and complex plots, the standard ROS GUI, \texttt{rqt}, was not
suitable, as it is not optimized for high-bandwidth display.
Instead, we developed a GUI based on \texttt{imgui}\footnote{\url{https://github.com/ocornut/imgui}}, an
immediate-mode GUI toolkit with OpenGL bindings. This allows us to decode and display the video streams
directly on the GPU. The GUI follows the \texttt{rqt} paradigm with windows that are arranged via
drag \& drop.

The most important monitoring display is shown in \cref{fig:sysmon}. Both operator station and avatar robot
run a \texttt{sysmon} node, which performs several checks with 1\,Hz.
These checks range from ``Is hardware device X connected?'' over ``Does component Y produce data?'' to ``Is the operator station properly calibrated?''.
The intention is simple: If all checks are successful, the support crew can start the run with confidence.
Indeed, our policy was that every time an undetected error or misconfiguration led to a sub-optimal test run,
a specific check for this condition was added.
Overall, checks are similar to unit tests in software engineering, but monitor the live system in hardware \textit{and} software.

\begin{figure}
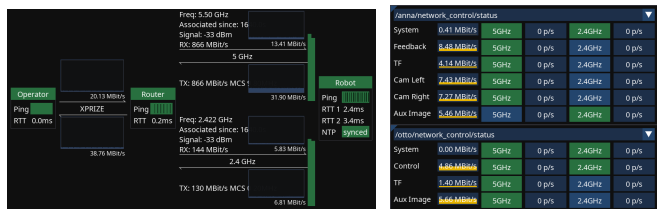

 \centering
 \nocompress[scale=0.31]{figures/imgui_network.png}\hfill\nocompress[scale=0.31]{figures/imgui_network_control.png}
 \caption{Network status. Left: Overview with individual network flows. Green boxes indicate hosts in the network.
 Right: Individual data groups can be configured to use 5\,GHz and/or 2.4\,GHz streams. The packet/s rates on the right
 indicate packet drops due to WiFi congestion.}
 \label{fig:network_control}
 \vspace{-1ex}
\end{figure}

The network subsystem is monitored and configured through two GUI components (see \cref{fig:network_control}).
An overview visualization shows bandwidths and parameters of each individual network connection. For the two
WiFi connections, signal strengths are also visualized as green bars.
A small control box allows switching data groups between the two WiFi connections. This ability allows quick
trouble-shooting by giving instant feedback on bandwidths and packet drops.

\begin{figure}
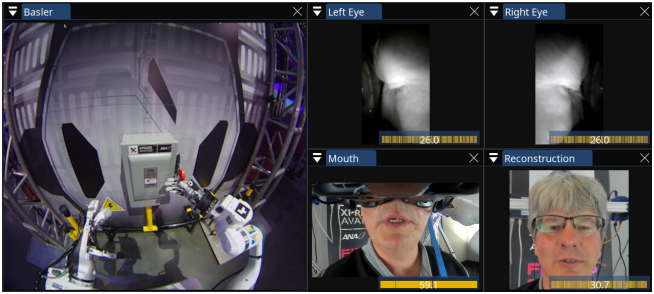

 \centering
 \nocompress[width=\linewidth]{figures/imgui_cams_new.jpg}
 \caption{Camera streams. Left: Raw wide-angle camera stream (left eye) from the robot.
 Right: Eye cameras, mouth camera, and reconstructed animated face of the operator.}\label{fig:cams}
\end{figure}

Finally, a section of the GUI with camera streams (see \cref{fig:cams}) together with headsets providing audio feedback give situational awareness to the
support crew.

\subsection{System Robustness}

Ensuring support crew situational awareness and the connectionless network system are features that make
the system more robust. However, there are many problems that can occur during a run, where manual
intervention is not possible without aborting the trial.
For this reason, we added auto-recovery mechanisms on multiple layers.

First, the Franka Emika Panda arms have independent safety systems which detect unsafe situations and
either perform a soft-stop (braking with motor power) or hard-stop (engaging hardware brakes and switching off motor power).
Since the operator can trigger both, e.g. by hitting an object with high speed, it is desirable
to recover from these conditions.
To this end, we modified the Panda firmware to be able to trigger recovery from an autonomous observer, which
restarts the arms as long as the manual E-Stop is not triggered.
During restart of the arm, the operator is shown a 3D model of the arm to indicate that the arm is restarting
and they should wait until the process is finished. The arm pose is then softly faded to the current operator pose
and operation can continue (see \cite{lenz2023bimanual} for more details).

Secondly, many hard- and software problems can be solved by simply restarting the affected processes~\citep{microreboot}.
As a simple example, restarting a device driver ROS node will recover after a transient disconnection of the device,
without the need to make the driver node itself robust against such events.
We stringently use the \texttt{respawn} feature of the ROS launch system to ensure that all nodes are
automatically restarted whenever they exit.
Watchdogs are integrated that force nodes which do not produce output to exit.

Finally, as a last line of defense, the main control PC is equipped with an external watchdog device.
Our software running on the control PC regularly resets this watchdog. Should the system hang completely
(which happened once during testing), the watchdog device will force a reset of the computer.
Consequently, the software is configured to auto-start again, automatically resuming operations.
The complete boot-and-recovery process takes less than one minute.

\section{Evaluation}\label{sec:eval}

\DeclareRobustCommand{\taskb}[1]{\strut\tikz{\node[fill=#1,rounded corners=1pt,minimum height=1.5ex,minimum width=1.5ex]{};}}
\begin{figure*}
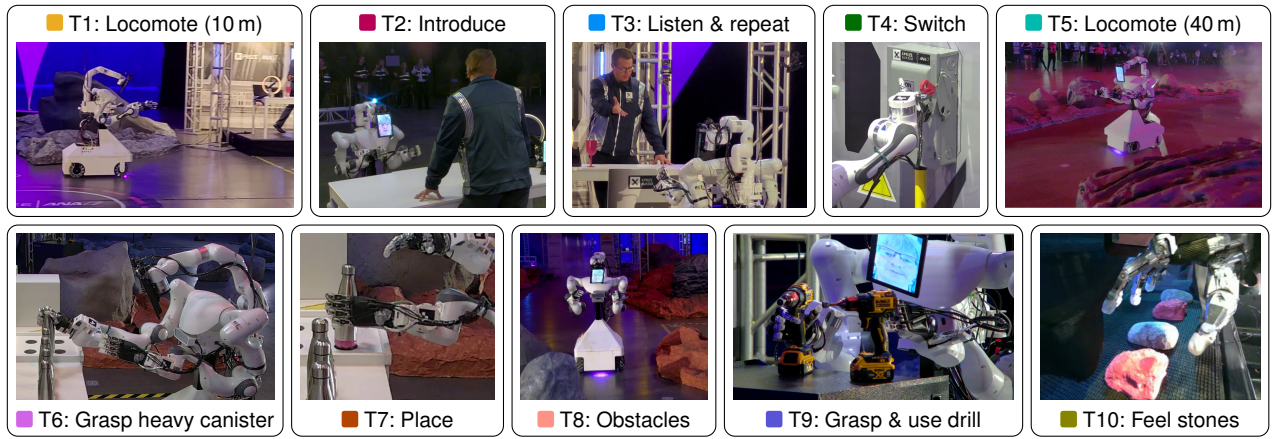

 \newlength{\theight}\setlength\theight{2.2cm}
 \centering
 \begin{tikzpicture}[
  t/.style={align=center,label distance=2pt, node distance=0.1cm, draw, rounded corners, text depth=0pt}
 ]
 \node[t] (t1) at (-8,0.3) {\taskb{t1} T1: Locomote (10\,m)\\[.4ex] \includegraphics[height=\theight,clip,trim=400px 350px 800px 300px]{figures/tasks/t1.png}};
  \node[t,right=of t1] (t2) {\taskb{t2} T2: Introduce\\[.4ex]\includegraphics[height=\theight,clip,trim=670px 230px 360px 200px]{figures/tasks/t2.png}};
  \node[t,right=of t2] (t3) {\taskb{t3} T3: Listen \& repeat \\[.4ex] \includegraphics[height=\theight,clip,trim=400px 500px 700px 0px]{figures/tasks/t3.png}};
  \node[t,right=of t3] (t4) {\taskb{t4} T4: Switch\\[.4ex] \includegraphics[height=\theight,clip,trim=1100px 270px 350px 280px]{figures/tasks/t4.png}};
  \node[t,right=of t4] (t5) {\taskb{t5} T5: Locomote (40\,m)\\[.4ex] \includegraphics[height=\theight,clip,trim=800px 230px 50px 160px]{figures/tasks/t5_oomph.png}};

  \node[t,anchor=north west,shift={(0,-0.1)}] (t6) at (t1.south west) {\includegraphics[height=\theight,clip,trim=1100px 450px 150px 200px]{figures/tasks/t6.png} \\[.4ex] \taskb{t6} T6: Grasp heavy canister};
  \node[t,right=of t6] (t7) {\includegraphics[height=\theight,clip,trim=1150px 500px 500px 350px]{figures/tasks/t7.png} \\[.4ex] \taskb{t7} T7: Place};
  \node[t,right=of t7] (t8) {\includegraphics[height=\theight,clip,trim=800px 240px 400px 200px]{figures/tasks/t8.png} \\[.4ex] \taskb{t8} T8: Obstacles};
  \node[t,right=of t8] (t9) {\includegraphics[height=\theight,clip,trim=700px 240px 50px 150px]{figures/tasks/t9_oomph.png} \\[.4ex] \taskb{t9} T9: Grasp \& use drill};
  \node[t,right=of t9] (t10) {\includegraphics[height=\theight,clip,trim=0px 0px 0px 0px]{figures/tasks/t10d.png} \\[.4ex] \taskb{t10} T10: Feel stones};
  \end{tikzpicture}
  \caption{Tasks of the ANA Avatar XPRIZE finals.
  \taskb{t1} T1: Short locomotion (approx. 10\,m) to the mission control desk.
  \taskb{t2} T2: The operator introduces themselves to the mission commander.
  \taskb{t3} T3: The operator receives mission details and confirms the tasks.
  \taskb{t4} T4: Activate the power switch.
  \taskb{t5} T5: Approx. 40\,m of locomotion.
  \taskb{t6} T6: Select a canister by weight (approx. 1.2\,kg).
  \taskb{t7} T7: Place the canister in the designated slot.
  \taskb{t8} T8: Navigate around obstacles.
  \taskb{t9} T9: Grasp and use the power drill to unscrew the hex bolt.
  \taskb{t10} T10: Select a rough-textured stone based on touch and retrieve it.
  }\label{fig:tasks}
  \vspace{-1ex}
\end{figure*}

\pgfplotstableread[col sep=comma]{data/task_timings.csv}{\loadedtable}
\pgfplotstabletranspose[colnames from=Team]\transposedtable\loadedtable
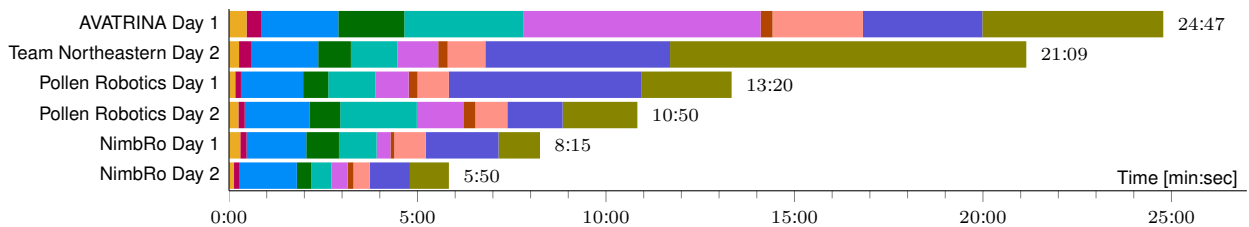
\begin{figure*}
 \centering
 \begin{tikzpicture}[
    font=\scriptsize\sffamily,
    ppin/.style={pin edge={latex-,red,thick},pin distance=8pt,font=\sffamily\scriptsize,draw=red,rounded corners},
  ]
  \begin{axis}[
      axis on top,
      anchor=north,
      height=4cm,
      clip=false,
      xmin=0,
      xmax=27,
      width=.85\linewidth,
      xbar stacked,
      bar width=10pt,
      axis x line*=left,
      axis y line*=left,
      minor x tick num=1,
      xtick align=outside,
      xtick pos=bottom,
      xtick distance=5,
      minor x tick num=4,
      ytick=data,
      ytick style={draw=none},
      yticklabels from table={\transposedtable}{colnames},
      ytick pos=left,
      x filter/.code={\pgfmathparse{#1/60}},
      xticklabel={ %
        \pgfmathsetmacro\hours{floor(\tick)}%
        \pgfmathsetmacro\minutes{(\tick-\hours)*0.6}%
        \pgfmathprintnumber{\hours}:\pgfmathprintnumber[fixed, fixed zerofill, skip 0.=true, dec sep={}]{\minutes}%
        },
      printtime/.style={%
        point meta=x,
        nodes near coords style={black},
        nodes near coords={
            \pgfkeys{/pgf/fpu=true}
            \pgfmathparse{\pgfplotspointmeta}
            \pgfmathsetmacro\hours{floor(\pgfplotspointmeta)}%
            \pgfmathsetmacro\minutes{(\pgfplotspointmeta-\hours)*0.6}%
            \pgfmathprintnumber{\hours}:\pgfmathprintnumber[fixed, fixed zerofill, skip 0.=true, dec sep={}]{\minutes}%
            \pgfkeys{/pgf/fpu=false}
        },
        nodes near coords align={horizontal}
      },
      cycle list={
        {draw=none,fill=t1},
        {draw=none,fill=t2},
        {draw=none,fill=t3},
        {draw=none,fill=t4},
        {draw=none,fill=t5},
        {draw=none,fill=t6},
        {draw=none,fill=t7},
        {draw=none,fill=t8},
        {draw=none,fill=t9},
        {draw=none,fill=t10}
      },
      every plot/.style={draw=none}
    ]

     \node[anchor=south east] at (axis cs:27.0,-0.7) {Time [min:sec]};

   \foreach \i in {1,...,9} {
    \addplot table [x index=\i, y expr=\coordindex] {\transposedtable};
   }
   \addplot+[printtime] table [x index=10, y expr=\coordindex] {\transposedtable};
  \end{axis}
 \end{tikzpicture}
 \vspace{-2.5ex}
 \caption{Per-task execution time for the top six competition runs solving all ten tasks in the ANA Avatar XPRIZE finals.}\label{fig:task_timing}
\end{figure*}

The ANA Avatar XPRIZE Finals took place in November 2022 in Long Beach, CA, USA.
After several down-selections over three years, 17 teams from 10 countries competed in the finals for a prize purse of \$8 million.
The developed avatar systems were evaluated by untrained operator and recipient judges in a series of ten tasks over one qualification and two competition days.
Only the top 16 teams and ties (qualification day) and top 12 teams (first competition day) advanced to the next day.
Both judges were selected from the international expert panel and unknown to the teams until 60\,min before the competition run.
Teams had 45\,min to train and familiarize the operator judge with their system.
The operator judge controlled the avatar robot located in the arena from the operator control room, out of range of direct visual or auditory feedback.
Information could only be exchanged between both locations via the avatar system.
The operator had up to 25\,min to complete all ten tasks (see \cref{fig:tasks}).
During the competition run, teams were not allowed to interact with the system or the judges.

\subsection{Analysis of Competition Scores}

Each competition run was scored based on task performance and judge experience.
\Cref{tab:final_points} shows the final scores of the top 12 teams.
One point was awarded for each task completed.
In addition, up to five points were given based on the experience of both judges.
Each judge scored up to one point per criterion if they felt the operator was present at the remote site and if they could clearly see and hear each other through the system.
The final point was given by the operator judge if the avatar system was easy and comfortable to use.
The maximum score of both competition days counted as the final result.
Ties were broken by completion time.

Four systems completed all ten tasks. Pollen Robotics' and our system were the only ones to solve all tasks on both days.
Most of the systems received 4.5 or the maximum 5 points for the judge experience.
Our team NimbRo won the competition with a perfect score of 15 points and the fastest completion time of 5:50\,min---almost twice as fast as the runner-up Pollen Robotics, who also received a perfect score with a time of 10:50\,min.

\begin{table}
 \centering
 \caption{Results of the ANA Avatar XPRIZE Finals}\label{tab:final_points}
 \begin{tabular}{rlrrrc}
  \toprule
  &  & \multicolumn{3}{c}{Points} & Time\\
  \cmidrule(lr){3-5} \cmidrule(lr){6-6}
   Rank & Team           & Total & Task & Judged & [mm:ss] \\
  \midrule
  \textbf{1} & \textbf{NimbRo}   & \textbf{15.0}  & \textbf{10}    & \textbf{5.0}   & \textbf{05:50}\\
  2 & Pollen Robotics                          & 15.0  & 10    & 5.0   & 10:50\\
  3 & Team Northeastern~\citep{luo2022towards} & 14.5  & 10    & 4.5   & 21:09\\
  4 & AVATRINA~\citep{marques2022commodity}    & 14.5  & 10    & 4.5   & 24:47\\
  5 & i-Botics~\citep{van2022comes}            & 14.0  & 9     & 5.0   & 25:00\\
  6 & Team UNIST                               & 13.5  & 9     & 4.5   & 25:00\\
  7 & Inbiodroid                               & 13.0  & 8     & 5.0   & 25:00\\
  8 & Team SNU~\citep{park2022team}            & 12.5  & 8     & 4.5   & 25:00\\
  9 & AlterEgo~\citep{lentini2019alter}        & 12.5  & 8     & 4.5   & 25:00\\
  10& Dragon Tree Labs                         & 11.0  & 7     & 4.0   & 25:00\\
  11& Avatar-Hubo~\citep{vaz2022immersive}     & 9.5   & 6     & 3.5   & 25:00\\
  12& Last Mile                                & 9.0   & 5     & 4.0   & 25:00\\
  \bottomrule
 \end{tabular}
 \vspace{-1ex}
\end{table}

\subsection{Task Completion Times}

We extracted the per task completion times for both competition days from the official video feed\footnote{\url{https://www.youtube.com/watch?v=lOnV1Go6Op0}} for the six runs completing all ten tasks (see \Cref{fig:task_timing} and \cref{tab:nimbro_task_time}).
Both of our competition runs were faster than any other successful run.
As our operator judge on Day~1 solved all tasks in 8:15\,min, giving us a comfortable lead,
our operator judge on Day~2 was instructed to take more risks.
In addition, we greatly increased our avatar's maximum base velocity for Day~2, resulting in much faster execution times
for all tasks involving larger locomotion (Tasks 1, 4, 5, and 8).
We encountered a minor network issue during Task~9 on Day~1 (see \cref{eval:failures}) which explains our longer execution time of 1:56\,min, compared to 1:04\,min on Day~2.
All remaining tasks (2, 3, 6, 7, and 10) were solved within the same time ($\pm$4\,sec.) on both days, showing the robustness of our system.

The shorter tasks \mbox{1-3} (locomotion and communication with the recipient judge)
and Task~7 (placing the canister into the designated slot) were consistently solved with similar execution times across the top six competition runs.
AVATRINA's system had a much slower drive compared to the top three teams, as evidenced by longer execution times for the locomotion tasks.
They also had problems during the manipulation in Task~6, which resulted in a software restart, costing 2:10\,min.
Pollen Robotics' longer locomotion time (Task~5) on Day~2 was similarly due to a reset of the operator control.
Larger differences in individual task execution times are due to subsystem failures or sub-optimal grasp poses in case of the drill (Task~9):
Both Pollen Robotics on Day~1 and Team Northeastern lost the first drill,
requiring grasping the second drill.
Team Northeastern then struggled to reach the stones in the box for Task~10,
maybe due to their kinematics with the wrist above the hand.
The left arm shut down completely due to collision with the top bar and could not recover.
However, the operator managed to retrieve the correct stone with the right arm after several attempts.

\begin{table}
 \centering\scriptsize\setlength\tabcolsep{1pt}
 \begin{threeparttable}
 \caption{Task Completion Times in ANA Avatar XPRIZE Finals}\label{tab:nimbro_task_time}
 \begin{tabular}{l@{\hspace{4pt}}rrrrrrrrrrr}
 \toprule
    Team / Run & T1 & T2 & T3 & T4 & T5 & T6 & T7 & T8 & T9 & T10 & Total\\
    \midrule
    Avatrina D1 & 0:28 & 0:23 & 2:03 & 1:45 & 3:10 & 6:17 & 0:19 & 2:24 & 3:10 & 4:48 & 24:47\\
    Northeastern D2 & 0:16 & 0:19 & 1:47 & 0:52 & 1:14 & 1:05 & 0:15 & 1:00 & 4:54 & 9:27 & 21:09\\
    Pollen Rob. D1 & 0:10 & \textbf{0:09} & 1:39 & 0:40 & 1:15 & 0:53 & 0:14 & 0:50 & 5:06 & 2:24 & 13:20\\
    Pollen Rob. D2 & 0:15 & \textbf{0:09} & 1:43 & 0:49 & 2:02 & 1:15 & 0:18 & 0:51 & 1:28 & 1:59 & 10:50\\
    NimbRo D1 & 0:18 & 0:10 & 1:35 & 0:52 & 1:00 & \textbf{0:22} & \textbf{0:06} & 0:50 & 1:56 & 1:06 & 8:15\\
    NimbRo D2 & \textbf{0:08} & \textbf{0:09} & \textbf{1:31} & \textbf{0:23} & \textbf{0:32} & 0:26 & 0:09 & \textbf{0:26} & \textbf{1:04} & \textbf{1:02} & \textbf{5:50}\\
    \midrule
    NimbRo D2-D1 & \color{teal}{-0:10} & \color{teal}{-0:01} & \color{teal}{-0:04} & \color{teal}{-0:29} & \color{teal}{-0:28} & \color{purple}{+0:04} & \color{purple}{+0:03} & \color{teal}{-0:24} & \color{teal}{-0:52} & \color{teal}{-0:04} & \color{teal}{-2:25}\\
    \bottomrule
 \end{tabular}
 We show times of the top six competition run in minutes. D1/D2: Day~1 / Day~2.
  \vspace{-1ex}
 \end{threeparttable}
\end{table}

Despite the small sample size, this time analysis shows the robustness and reliability of our system.
Comparing our completion times with the rest of the competition suggests that our system is easier and faster to use and provides sufficient intuitive situational awareness to the operator.

\subsection{System Failures \& Recovery}\label{eval:failures}

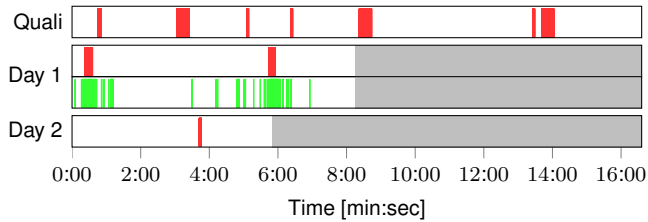
\begin{figure}
 \centering
 \begin{tikzpicture}
  \begin{groupplot}[
    group style={rows=4, vertical sep=0pt, ylabels at=edge bottom, x descriptions at=edge bottom},
    axis on top,
    ybar,
    ymin=-0.03,
    ymax=1.1,
    xmin=0,
    xmax=16.6,
    xtick=\empty,
    xtick pos=bottom,
    xlabel={Time [min:sec]},
    width=1.06\linewidth,
    height=2cm,
    ytick=\empty,
    /pgf/bar width=0.333,
    enlarge y limits=false,
    x filter/.code={\pgfmathparse{#1/60}},
    xticklabel={ %
      \pgfmathsetmacro\hours{floor(\tick)}%
      \pgfmathsetmacro\minutes{(\tick-\hours)*0.6}%
      \pgfmathprintnumber{\hours}:\pgfmathprintnumber[fixed, fixed zerofill, skip 0.=true, dec sep={}]{\minutes}%
      },
    cmd/.style={draw=red!80,fill=red!50,/pgf/bar width=1},
    fb/.style={draw=blue!80,fill=blue!50},
    vision/.style={draw=green!80,fill=green!50,/pgf/bar width=.01}
  ]

  \nextgroupplot
  \addplot[cmd, /pgf/bar width=0.002] table [x=time, y expr=1] {data/run_anna_2022-11-04-05-31-38.onlyrun.command_fail.txt};
  \coordinate (q1) at (axis description cs:0,0.5);

  \coordinate (q2) at (axis description cs:0,0.5);

  \nextgroupplot[yshift=-0.1cm]
  \addplot[cmd, /pgf/bar width=0.002] table [x=time, y expr=1] {data/run_anna_2022-11-05-03-13-40.onlyrun.command_fail.txt};
  \coordinate (d11) at (axis description cs:0,0.5);
  \draw[draw=none,fill=gray!50] (axis cs:8.25,2) rectangle (axis cs:17,-0.1);

  \nextgroupplot
  \addplot[vision] table [x=time, y expr=1] {data/run_otto_2022-11-04-19-13-40.vronly.onlyrun.vision_errors.txt};
  \coordinate (d12) at (axis description cs:0,0.5);
  \draw[draw=none,fill=gray!50] (axis cs:8.25,2) rectangle (axis cs:17,-0.1);

  \nextgroupplot[yshift=-0.1cm,xtick={}]
  \addplot[cmd, /pgf/bar width=0.002] table [x=time, y expr=1] {data/run_anna_2022-11-05-23-55-24.runonly.command_fail.txt};
  \coordinate (d21) at (axis description cs:0,0.5);
  \draw[draw=none,fill=gray!50] (axis cs:5.8333,2) rectangle (axis cs:17,-0.1);

  \coordinate (d22) at (axis description cs:0,0.5);

  \end{groupplot}

  \node[anchor=east] at ($(q1)!0.5!(q2)$) {Quali};
  \node[anchor=east] at ($(d11)!0.5!(d12)$) {Day 1};
  \node[anchor=east] at ($(d21)!0.5!(d22)$) {Day 2};
 \end{tikzpicture}
 \caption{Network latency events during finals. 
Delays over 100\,ms in the arm command channel (red), temporarily put the arms in pause mode.
 From Day 1 on, the arm commands were transmitted redundantly.
 Additional logs from Day 1 show video decoding errors resulting from packet loss (green).
 }
 \label{fig:latency}
\end{figure}

During all three competition runs, the arm recovery system was put to the test.
On qualification day, we had switched off most traffic on the 2.4\,GHz band since it had proven unstable in the team garages
due to channel congestion. During the run, there were certain intervals with higher packet jitter on the 5\,GHz band.
Since redundancy on the arm commands was unavailable, the arm controllers on the robot
disabled themselves when not receiving commands (see \cref{fig:latency}). Thankfully, the packet jitter was only transient, so that auto-restarting the controllers was possible.
Additionally, during the final task, the left wrist hit the boundary of the stone box with enough force to disable the arm. 
Pressure on the arm prevented immediate auto-recovery but the operator resolved the situation by lowering the torso, reducing the pressure and allowing the arm to restart.

After experiencing the packet jitter on qualification day, we enabled redundancy for the arm commands, which greatly reduced this
error class from seven instances in qualification to two on Day~1 and only one on Day~2.
This demonstrates the utility of the redundant communication system.

Arm shutdowns due to excessive force happened once on Day~1 again in the stone box, and on Day~2
while putting a canister back on the table. In both cases, auto-recovery immediately succeeded and the run could continue without problems.

\subsection{Lessons Learned}

\subsubsection{Robustness}
Despite the extensive experience available in our group, this is the most complex system we ever built.
Good monitoring, failure tolerance, and auto-recovery were extremely important for success.
In contrast to other teams, technical problems did not cause major delays or failures.

\subsubsection{Frequent testing under competition conditions}
Having a high test frequency allowed us to identify and address several issues that were annoying or uncomfortable
for operators, e.g. resulting in the improved rudder device.
Frequent tests also helped the support crew to establish routine in efficiently training the operators.

\subsubsection{1:1 correspondence is best}
The connection between operator and avatar needs to be as close to identity as possible.
Avoiding any scaling, offsetting or 3D processing helps operators to quickly
immerse into the system. In particular, correct hand-eye transformations let operators identify the avatar's hands as their own.

\section{Conclusion}

We presented the extended and updated NimbRo avatar system, which won the ANA Avatar XPRIZE finals.
Key improvements, compared to the semifinals system~\cite{schwarz2021nimbro}, such as a new base design, a linear actuator to adjust the torso height, haptic perception, monitoring tools, failure tolerance, and robust wireless communication enabled this success.

\printbibliography

\end{document}